\documentclass[twocolumn]{antresearch}

\usepackage{amsmath}
\usepackage{amssymb}
\usepackage{balance}
\usepackage[ruled,vlined,linesnumbered]{algorithm2e}
\usepackage{float}
\usepackage{tikz}
\usetikzlibrary{shapes,arrows,arrows.meta,positioning,fit,backgrounds,calc}
\graphicspath{{figures/}}
\providecommand{\Description}[1]{}
\title{Know When to Stop, Where to Restart:\texorpdfstring{\\}{ }Accelerating Multi-Turn Agentic On-Policy Distillation}

\author[1,2,\ddagger,*]{Zhiyu Gui}
\author[1,\ddagger]{Kexin Huang}
\author[2,\dagger]{Jia Guo}
\author[1]{Junkang Wu}
\author[2]{Zihao Wang}
\author[2]{Zhiqiang Zhang}
\author[2]{Jun~Zhou}
\author[1]{Jiancan Wu}
\author[1,\dagger]{Xiang Wang}
\affiliation[1]{University of Science and Technology of China}
\affiliation[2]{Ant Group}
\contribution[\ddagger]{Equal contribution}
\contribution[\dagger]{Corresponding authors}
\contribution[*]{Work done during internship at Ant Group.}

\date{August 2026}

\abstract{
On-policy distillation (OPD) has become a standard approach for transferring
capabilities from large teachers to compact students. Its cost, however, is
dominated by autoregressive student rollouts and scales poorly in multi-turn
agentic settings. Existing acceleration methods truncate or relocate the
supervision signal according to fixed, offline budgets, despite substantial
variation in teacher-signal reliability both within and across trajectories.
Our empirical analysis on $\tau^2$-bench reveals a clear structure in this
variation: informative supervision is concentrated in the prefix of each
turn, and---most importantly for multi-turn
agentic training---the cross-turn loss of teacher endorsement is temporally
locked to the student's first erroneous action rather than accumulating
gradually over turns. Building on these findings, we propose \textbf{STRIDE}
(\textbf{ST}op-and-\textbf{R}estart on-pol\textbf{I}cy \textbf{D}istillation
acc\textbf{E}leration), which combines two complementary techniques:
\emph{adaptive early stopping}, which terminates a rollout
once the cumulative teacher log-probability falls below an
out-of-distribution threshold, and a \emph{prefix buffer}, which caches
high-quality prefixes and restarts generation at the weakest correct turn.
Together, these mechanisms induce a data-driven curriculum that progressively
extends coverage to later turns. On $\tau^2$-bench
retail, our method matches full-trajectory OPD and exceeds the 30B
teacher at a $3.73\times$ speedup, surpasses the baseline itself at
$2.34\times$, and retains a $4.51\times$ speedup under
  cross-domain multi-teacher training. As a supplementary generalization test
  beyond the agentic setting, STRIDE outperforms full OPD on AIME~2025 at a
  $5.10\times$ speedup and on AIME~2024 at a $3.08\times$ speedup; averaged
  across the two evaluations, both fixed-budget truncation baselines remain
  below full OPD.
}

\gtechdata[Keywords]{On-policy distillation; agentic training; efficient post-training}

\begin{document}
\maketitle

\begin{figure*}[t]
  \centering
  \includegraphics[width=\textwidth]{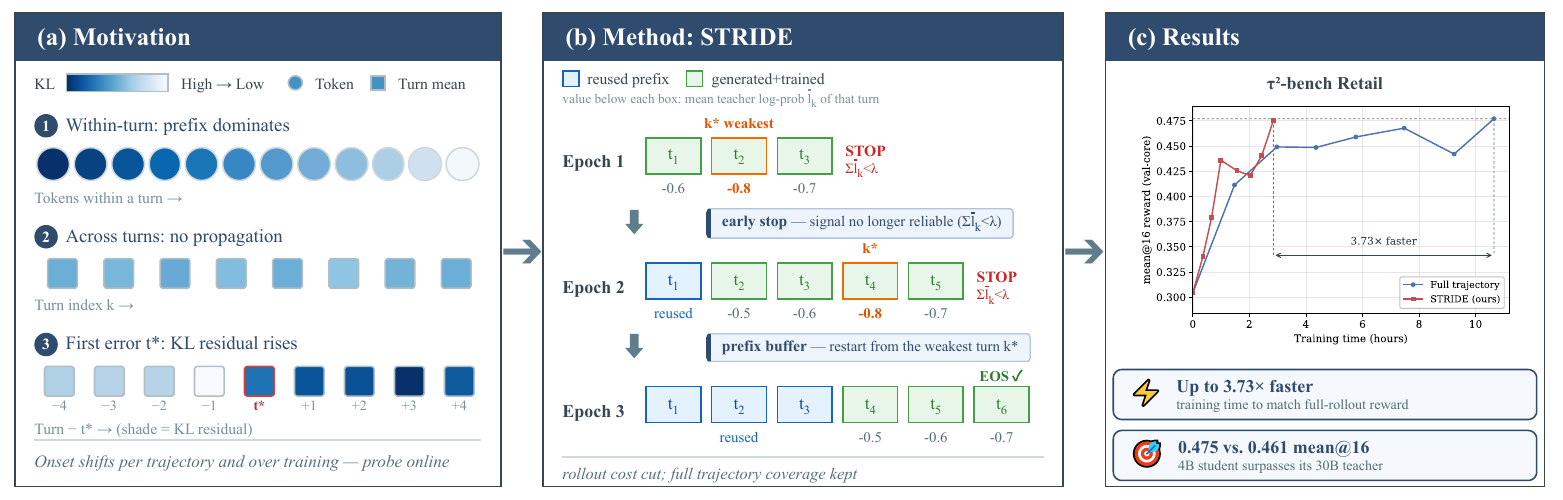}
  \Description{Three-panel overview: (a) where the teacher signal remains
  reliable in agentic on-policy distillation, (b) the STRIDE acceleration
  framework, and (c) its speed--accuracy results.}
  \caption{Overview of STRIDE. \textbf{(a)}~Where the teacher signal
           remains reliable in agentic OPD (Observations~1--3,
           \S\ref{sec:ood-analysis}).
           \textbf{(b)}~STRIDE on one task across training epochs: adaptive
           early stopping terminates the rollout once the cumulative teacher
           log-probability crosses $\lambda$, and the prefix buffer caches the
           correct prefix and restarts the next rollout from the weakest
           correct turn $k^{*}$, which shifts rightward as the student
           improves (per-turn values are illustrative). \textbf{(c)}~On
           $\tau^2$-bench retail, STRIDE matches full-trajectory
           OPD and exceeds the 30B teacher at a $3.73\times$ speedup.}
  \label{fig:overview}
\end{figure*}

\section{Introduction}
\label{sec:intro}

Large language model (LLM) agents that interleave reasoning, tool use, and
environment interaction over many turns are increasingly deployed in
practice~\cite{openai2025codex,anthropic2025claudecode},
yet serving a large agent model at scale remains costly. Knowledge
distillation offers a standard remedy~\cite{hinton2015distilling}: train a
compact student to imitate a
stronger teacher. Classic off-policy distillation, however, trains the student
on teacher-generated data, whose distribution diverges from the states the
student actually visits at deployment; the mismatch compounds over
autoregressive generation and is amplified across agentic interactions~\cite{liao2026multiturnonpolicydistillationprefix,zhong2026sodstepwiseonpolicydistillation}.
On-policy distillation (OPD) removes the mismatch at its source: the student
generates its own trajectories and the teacher supplies dense token-level
feedback on them. OPD now underpins industrial training
recipes~\cite{qwen2025qwen3,deepseek2026v4,lu2025thinkingmachines}.

Removing this mismatch, however, comes at a price: online data generation.
Unlike off-policy distillation, which trains on a fixed teacher-generated
dataset that is collected once and reused throughout training, OPD must
repeatedly roll out the evolving student to collect fresh on-policy
trajectories---an overhead that is especially pronounced in multi-turn
agentic settings.
Existing acceleration methods reduce this cost through static decisions
made offline, before or outside training: truncating supervision to a fixed
prefix, which directly shortens the
rollout~\cite{zhang2026fastopd,ji2025upft}; or moving OPD offline entirely
by sampling rollouts once and precomputing teacher scores on them, which
removes both the online rollout and the live teacher but freezes the training
distribution at the initial policy~\cite{wu2026lightning}. Such static
choices are misaligned with how the reliability of the distillation signal
actually behaves:
\begin{itemize}
  \item Within a response, the corrective signal generally concentrates near
        the beginning, but the rate at which it decays---and thus the length
        of the useful prefix---varies substantially across responses;
  \item Within a multi-turn trajectory, the teacher's endorsement can
        collapse abruptly once the student takes a single wrong action, at a
        depth that differs from trajectory to trajectory and that itself
        shifts as the student improves over training.
\end{itemize}
As our experiments confirm (\S\ref{sec:single-teacher},
\S\ref{sec:math}), a static rule therefore inevitably over-truncates some
trajectories---discarding valid supervision---while under-truncating
others---retaining tokens whose teacher signal is no longer meaningful.

This structure of the signal explains why offline remedies can fail: the
failure point differs across trajectories and shifts over the course of
training, so the reliability of the teacher signal must be assessed
\emph{during} the rollout itself. Our analysis on the $\tau^2$-bench retail
domain (\S\ref{sec:sparsity}; Figure~\ref{fig:overview}a) confirms this:
supervision concentrates in the within-turn prefix, the fade does not
propagate across turns, and the cross-turn loss of teacher endorsement is
temporally locked to the student's first erroneous action. The sharp,
action-locked onset of failure can therefore be caught by an online probe
as it happens---enabling adaptive early stopping---while everything before
it remains valid supervision that need not be regenerated---enabling safe
prefix reuse. Both mechanisms are driven by a single statistic that OPD
already computes during the rollout: the per-turn mean teacher
log-probability.

Taken together, these findings call for acceleration that is both
\emph{adaptive} and \emph{online}. We propose \textbf{STRIDE} (Stop-and-Restart on-policy
Distillation acceleration), which combines two complementary techniques
(\S\ref{sec:method}; overview in Figure~\ref{fig:overview}). \emph{Adaptive early stopping} terminates a rollout as
soon as the cumulative teacher log-probability crosses a threshold, directly
cutting the dominant rollout cost while responding to the action-locked onset
of out-of-distribution (OOD)
drift rather than penalizing long-but-valid trajectories. Because pure
truncation forfeits coverage of the later turns, a \emph{prefix buffer}
caches high-quality trajectory prefixes, restarts the next rollout of the same
task from the weakest correct turn, and thereby slides the training window
rightward as the student improves---a data-driven curriculum that requires no
hand-designed schedule.

Our experiments focus on two multi-turn agentic settings---single-teacher OPD
on $\tau^2$-bench retail and cross-domain multi-teacher OPD on retail and
telecom---supplemented by a single-turn math-reasoning study on AIME that
tests whether the method's effectiveness depends exclusively on the
multi-turn structure
(\S\ref{sec:experiments}). On retail, STRIDE
matches the full-trajectory baseline and exceeds the 30B teacher---while
attaining the best pass@16 overall---at a $3.73\times$ speedup, and surpasses
the baseline by $+1.3\%$ mean@16 at $2.34\times$; multi-teacher training
  retains a $4.5\times$ speedup, and the method outperforms full OPD on
  AIME~2025 at $5.10\times$ and on AIME~2024 at $3.08\times$. In contrast,
  fixed-budget truncation remains below full OPD when averaged across the two
  AIME evaluations.

Our contributions are as follows:
\begin{itemize}
  \item An empirical analysis of teacher-signal reliability in agentic OPD
        that localizes the failure of the teacher signal: supervision
        concentrates in the within-turn prefix, and the cross-turn loss of
        endorsement is locked to the first erroneous action rather than to
        turn index or trajectory length (\S\ref{sec:sparsity}).
  \item \emph{Adaptive early stopping}, an online truncation rule driven by
        the cumulative teacher log-probability, which detects the onset of
        OOD drift instead of imposing a fixed budget
        (\S\ref{sec:early-stopping}).
  \item A \emph{prefix buffer} that couples prefix reuse with weakest-turn
        localization, recovering the coverage lost to truncation through a
        data-driven curriculum over trajectory depth
        (\S\ref{sec:prefix-buffer}).
  \item A comprehensive evaluation across single-teacher and multi-teacher
        agentic OPD, plus a supplementary single-turn math-reasoning study,
        characterizing the speed--quality Pareto frontier and showing that
        acceleration does not compromise the student's ability to exceed its
        teacher (\S\ref{sec:experiments}).
\end{itemize}

\section{Related Work}
\label{sec:related}

\textbf{On-policy distillation.} Off-policy distillation suffers from a
distribution mismatch between the teacher's training data and the student's
deployment distribution, an error that compounds over autoregressive
generation. On-policy distillation (OPD) lets the student generate its own
training data and uses the teacher only to score it, eliminating the
mismatch at the source~\cite{fang2026rubricbasedonpolicydistillation,sun2026simctrecoveringlostsupervision}. GKD~\cite{agarwal2024gkd} introduced systematic
  on-policy sampling into LLM distillation, MiniLLM~\cite{gu2024minillm} cast
  reverse-KL OPD as policy-gradient optimization, and
  \citet{yang2026gopd} proved that OPD is a special case of densely
KL-constrained RL whose reward extrapolation lets the student exceed the
teacher---a property our agentic results confirm. OPD has since become a
production staple~\cite{sun2026easyopdeasytouseonpolicydistillation}: Qwen3~\cite{qwen2025qwen3} formalized a two-stage recipe
of off-policy cold start followed by on-policy distillation, and Thinking
Machines~\cite{lu2025thinkingmachines} reported RL-grade quality at roughly
a tenth of RL cost by replacing sparse RL rewards with dense per-token
teacher feedback.

  \textbf{Teacher-signal reliability and OOD drift.} A line of work examines
  when the teacher signal itself becomes unreliable.
  \citet{fu2026revisiting} show that once the student prefix leaves the
  teacher's typical support the teacher's conditional distribution is no longer
  reliable; \citet{liu2026prefixteach} identify local teachability
  collapse in strong-to-weak OPD, motivating prefix-only supervision;
  \citet{liu2026sfd} formalize the within-turn aspect of this phenomenon
  as \emph{supervision fidelity decay} and counter it with a lookahead reward;
  and \citet{li2026posconf} restrict confidence-based decisions to
reliably calibrated position intervals. TIP~\cite{tip2026} shows that
training on a fraction of tokens can match full-token OPD, but its selection
presupposes a complete rollout and a full teacher pass, so it reduces
neither the rollout nor the teacher cost. Our analysis in
§\ref{sec:ood-analysis} is complementary: it establishes the within-turn
fade empirically, shows that it does \emph{not} transfer across turns, and
localizes the cross-turn OOD trigger to a specific erroneous action rather
than to turn index or trajectory length.

\textbf{Accelerating OPD.} Since the cost of OPD is dominated by the student
rollout, several methods strip or relocate the supervision signal at a fixed
budget. Fast~OPD~\cite{zhang2026fastopd} truncates supervision to a fixed
student prefix, and Lightning~OPD~\cite{wu2026lightning} moves OPD offline
entirely at the price of freezing the training distribution at the initial
policy. By contrast, our early stopping is adaptive and online, and the
prefix buffer recovers the coverage that truncation forgoes. Closest to our
work are four concurrent methods: Prune-OPD~\cite{yang2026pruneopd}
truncates drifted rollouts in real time by token-level student--teacher
compatibility; TurnOPD~\cite{zhou2026turnopd} budgets the rollout depth of
multi-turn agent training from probe-based turn statistics;
TCOD~\cite{wang2026tcod} controls the exposed trajectory depth by a manually
designed temporal-curriculum schedule and regenerates every rollout from
scratch; and
ReOPD~\cite{liao2026multiturnonpolicydistillationprefix} replays
pre-collected \emph{teacher} prefixes offline: the student is on-policy only
at the single supervised step, and reliability is reduced to a static
position-decaying schedule over a frozen prefix pool---whereas
Observation~3 shows the failure point is action-locked and shifts over
training. All four allocate computation by supervision reliability, as we
do, but none recovers what truncation removes.

\section{Teacher-Signal Reliability and the Cause of OOD Drift}
\label{sec:sparsity}

\subsection{Preliminaries: Token-Level OPD}
\label{sec:sparsity-prelim}

We begin by fixing the minimal notation used throughout. In on-policy
distillation (OPD), the student model $\pi_\theta$
generates a response $y \sim \pi_\theta(\cdot \mid x)$ from an input $x$, and
the teacher model $\pi_T$ computes log-probabilities at each student-generated
token, supplying a dense token-level supervision signal. OPD is theoretically
grounded in minimizing the sequence-level reverse KL divergence between the
student and teacher policies~\cite{gu2024minillm}. Applying the
policy-gradient theorem to this objective yields a gradient estimator in
which the update at position $t$ couples the immediate reward at $t$ with all
future rewards:
\begin{equation}
  \begin{split}
    g_t^{\text{seq}}
    &= \nabla_\theta \log\pi_\theta(y_t\mid y_{<t})
       \cdot \sum_{t'=t}^{|y|} r_{t'},\\
    r_t &= \log\pi_T(y_t\mid y_{<t}) - \log\pi_\theta(y_t\mid y_{<t}),
  \end{split}
  \label{eq:seq-grad}
\end{equation}
where $g_t^{\text{seq}}$ denotes the gradient contribution of position $t$
under the sequence-level objective, and $r_t$ is the per-token
teacher--student log-probability ratio. This
estimator is unbiased, but its variance grows rapidly with sequence length
(up to $O(T^4)$ under bounded rewards and
gradients~\cite{li2026rethinking}), which is prohibitive
for long agentic trajectories. Token-level OPD, which we adopt throughout,
removes the future-reward coupling---equivalently, setting the discount
factor to zero---so that each position is updated using only its local
reward:
\begin{equation}
  \begin{aligned}
  \mathcal{L}_{\text{OPD}}
  = \mathbb{E}_{x,\, y\sim\pi_\theta}\!\Bigl[
      &-\sum_t \log\pi_\theta(y_t\mid y_{<t}) \cdot r_t \\
      &\cdot \mathbf{1}_{\text{trainable}}(t)
    \Bigr],
  \end{aligned}
  \label{eq:opd-loss}
\end{equation}
where $r_t$ is treated as a constant (stop-gradient), and the indicator
$\mathbf{1}_{\text{trainable}}(t)$ masks out tokens returned by the
environment as observations, ensuring that only tokens the student generates
on its own participate in the loss. Token-level OPD is a biased approximation
of the sequence-level objective, but it reduces the worst-case gradient
variance from $O(T^4)$ to $O(T^2)$, making it a practical choice for
long-horizon training.

In agentic settings, a trajectory $\tau$ consists of alternating student
actions and environment observations:
\begin{equation}
  \tau = [x,\, a_0^{\text{think}},\, a_0^{\text{tool}},\, o_1,\,
          a_1^{\text{think}},\, a_1^{\text{tool}},\, o_2,\, \ldots]
\end{equation}
We partition the student's generation by interaction turn: turn $k$ comprises
the reasoning block and tool call that the student produces before the
$(k{+}1)$-th environment observation. Let
$y^{(k)} = (y^{(k)}_1, \ldots, y^{(k)}_{n_k})$ denote the token sequence
generated in turn $k$, with $n_k$ tokens, and $h^{(k)}$ the history context up
to turn $k$. The per-turn mean teacher log-probability is
\begin{equation}
  \bar{l}_k = \frac{1}{n_k}\sum_{i=1}^{n_k}
    \log\pi_T\!\left(y^{(k)}_i \,\middle|\, h^{(k)}, y^{(k)}_{<i}\right).
  \label{eq:turn-logp}
\end{equation}
Intuitively, $\bar{l}_k$ characterizes the average endorsement the teacher
  grants to the student's turn-$k$ output: higher values indicate that the
  student's generation lies closer to the teacher's support.
  \citet{fu2026revisiting} have shown that once the teacher's conditional
distribution $\pi_T(\cdot\mid y_{<t})$ over a prefix falls into an
out-of-distribution (OOD) region it is no longer reliable, and the resulting
log-probability signal $\log\pi_T(y_t\mid y_{<t})$ may produce misleading
gradient directions. $\bar{l}_k$ and its evolution across turns therefore
serve both as the object of the empirical analysis that follows
(§\ref{sec:ood-analysis}) and as the core online probe underlying the
early-stopping criterion (§\ref{sec:early-stopping}).

\subsection{Empirical Analysis of Teacher-Signal Reliability}
\label{sec:ood-analysis}

We examine where the teacher signal remains reliable and, more importantly,
why it breaks down. The analysis is conducted on the
$\tau^2$-bench retail domain ($456$ tasks, $5{,}856$ turns) with student
Qwen3-4B and the Qwen3-30B-A3B-Thinking-2507 teacher used in our
single-teacher experiments (§\ref{sec:single-teacher}). For each turn
we track the mean teacher log-probability $\bar{l}_k$
(Eq.~\ref{eq:turn-logp}) and the mean per-token student--teacher KL
(estimated per token by $\log\pi_\theta(y_t\mid y_{<t}) - \log\pi_T(y_t\mid
y_{<t})$, i.e.\ $-r_t$ in Eq.~\ref{eq:seq-grad}).

\begin{figure}[t]
  \centering
  \includegraphics[width=\columnwidth]{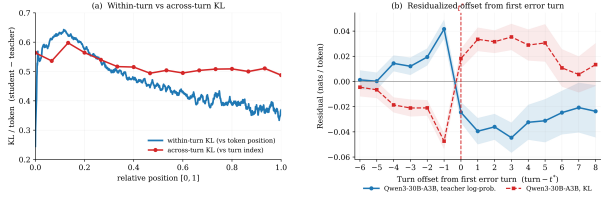}
  \Description{Per-token student--teacher KL divergence versus the relative
  position within a turn and versus the turn index, showing that the teacher
  signal fades within a turn and collapses after the first erroneous action.}
  \caption{\textbf{(a)}~Per-token student--teacher KL versus the relative
           position within a turn (blue) and versus the turn index (red) on
           $\tau^2$-bench retail (student Qwen3-4B, teacher
           Qwen3-30B-A3B-Thinking-2507). \textbf{(b)}~Residualized
           teacher log-probability (solid) and student--teacher KL (dashed)
           versus the offset from the first-error turn $t^{*}$; shaded bands
           indicate $\pm1$ standard error.}
  \label{fig:ood-cause}
\end{figure}

\textbf{Observation 1 (within-turn).} The within-turn profile is flat only at
the very opening, where the response-framing tokens dominate: past the first
${\sim}150$ tokens, the per-token student--teacher KL peaks at $+0.62$ and
then declines monotonically across the remaining deciles to $+0.37$, while
the mean teacher log-probability rises steadily from $-0.78$ to $-0.65$ over
the same span (Figure~\ref{fig:ood-cause}a). The student--teacher gap thus
narrows as generation proceeds---the teacher grants increasing endorsement to
the student's own tokens, and its scores progressively lose the power to
discriminate correct from incorrect continuations.
Effective supervision is therefore concentrated in the response prefix, the
empirical basis of prefix-only distillation~\cite{zhang2026fastopd,
  liu2026prefixteach}. The fade mirrors two independent findings:
  \citet{liu2026sfd} show that the teacher's distribution loses confidence and
  discriminative power as the prefix grows, while \citet{li2026posconf} report
  late-response inverse calibration of OPD confidence.

\textbf{Observation 2 (across turns).} The within-turn fade does not propagate
to the turn axis. Were the teacher to assimilate the student distribution by
reading an ever-growing body of student text turn after turn, $\bar{l}_k$
would rise monotonically with $k$. It does not: across turns $0$--$12$, the
mean teacher log-probability is essentially flat ($+0.07$) and the KL shift
is correspondingly small ($-0.06$; Figure~\ref{fig:ood-cause}a), both far
weaker than the within-turn shift and non-monotone. The reason is structural rather than statistical: the
Qwen3~\cite{qwen2025qwen3} chat template drops the reasoning content of
previous assistant turns when composing later-turn prompts, so the teacher
observes too little student text to fit its distribution turn by turn. Any pronounced
cross-turn decline of $\bar{l}_k$ therefore cannot be attributed to progressive
assimilation and must have another origin.

\textbf{Observation 3 (origin of cross-turn drift).} For each failed task we
identify the first error turn $t^{*}$---the earliest turn at which the agent's
action departs from the environment's reference behavior---and align turns by
the offset $\mathrm{turn}-t^{*}$. To isolate structure independent of turn
position we residualize against the per-turn-index mean, removing exactly the
weak trend documented in Observation~2. As Figure~\ref{fig:ood-cause}b shows,
the teacher-log-probability residual is flat and mildly positive before
$t^{*}$, turns over at $t^{*}$, and remains negative thereafter, while the KL
residual mirrors it. The residualized before-versus-after effect (Cohen's
$d$, computed over the $267$ failed tasks for which $t^{*}$ is identified) is
$+0.30$ on $\bar{l}$ and $-0.33$ on the KL. The loss of endorsement is thus
temporally locked to the erroneous action rather than to turn index or
trajectory length: a wrong action in turn $t^{*}$ drives the subsequent state
off the teacher's support---the single-turn trigger of the inter-turn error
accumulation described by TCOD~\cite{wang2026tcod}.

\textbf{Implications.} The reliable signal occupies precisely the region the
teacher can still support: within a turn, the high-contrast prefix
(Observation~1); across turns, the segment preceding an erroneous
action (Observation~3). In both regimes the boundary of this region is tracked
by $\bar{l}_k$. This calls for an online detector rather than a post-hoc
selection of ``critical tokens,'' which would presuppose the full
rollout~\cite{tip2026}:
detect the crossing of the support boundary through the cumulative decline of
$\bar{l}_k$ and truncate there (§\ref{sec:early-stopping}), while a prefix
buffer recovers the coverage that truncation forgoes on later-stage behavior
(§\ref{sec:prefix-buffer}).

\section{Method}
\label{sec:method}

\subsection{Early Stopping via Teacher Log-Probability}
\label{sec:early-stopping}

\emph{Motivation.} §\ref{sec:ood-analysis} establishes that the per-turn
teacher log-probability $\bar{l}_k$ tracks the reliability of the teacher's
supervision, and that the cross-turn loss of endorsement is sharp and locked
to the first erroneous action (Observations~2 and~3)---precisely the kind of
drop an online probe can catch as it happens. Tokens generated after such a
drop lie off the teacher's support and carry no reliable gradient. The
natural response is to terminate the rollout \emph{as soon as} the
cumulative teacher log-probability signals that the trajectory has crossed
into an OOD region, rather than to keep computing over an unreliable one.

\emph{Detection criterion.} We use the per-turn mean teacher log-probability
$\bar{l}_k$ (Eq.~\ref{eq:turn-logp}) as an online OOD probe. During normal
interaction $\bar{l}_k$ typically resides in a moderately negative range,
reflecting the teacher's partial endorsement of the student's plausible but
imperfect outputs; when the student drifts into an OOD region, $\bar{l}_k$
drops sharply, rapidly pushing the cumulative sum past the threshold.
Formally, we terminate the rollout at the first turn $K^*$ for which
\begin{equation}
  \sum_{k=1}^{K^*} \bar{l}_k \;<\; \lambda,
  \label{eq:stop-rule}
\end{equation}
where the threshold $\lambda$ controls OOD sensitivity. Generation stops at
turn $K^*$, but tokens from turns $1$ through $K^*$---including the
triggering turn---participate in the OPD loss: under token-level OPD the
teacher's low log-probability on the erroneous action translates into a
negative per-token reward, so the triggering turn supplies an explicit
negative signal that teaches the student which action to avoid, and
discarding it would forfeit precisely the most informative corrective
signal. Because the cross-turn decline of $\bar{l}_k$ is locked to an
erroneous action rather than to turn index or trajectory length
(Observation~3), this rule responds to erroneous states instead of
penalizing long-but-valid trajectories.

\emph{Acceleration mechanism.} Early stopping directly reduces the dominant
cost of OPD: the student's autoregressive rollout, which accounts for the bulk
of per-step training time in agentic settings---every discarded turn is
generation that never happens. Terminating early also avoids teacher scoring
and gradient computation over the truncated suffix, but these savings are
secondary: a teacher forward pass over the retained prefix is cheap relative
to autoregressive generation, so the measured speedup is driven essentially by
the shortened rollout.

\emph{Synergy with the prefix buffer.} When the prefix buffer is enabled, the
early-stopping cumulative sum is taken only over the newly generated turns of
the current rollout, i.e.\ $\sum_{j=K_0+1}^{K} \bar{l}_j < \lambda$, where
$K_0$ is the number of turns already covered by the prefix. This design
keeps historical turns reused from the prefix out of the stopping
decision: their accumulated teacher signal would otherwise drown out the
signal of the newly generated turns and blunt the probe's OOD-detection
sensitivity.

\emph{The $\lambda$ control surface.} A smaller $|\lambda|$ gives an
aggressive stopping policy---high OOD sensitivity and maximum speedup, at the
risk of misjudging the boundary region and of insufficient coverage; a larger
$|\lambda|$ is more conservative, tolerating more of the boundary-signal
region at the cost of a reduced speedup. The appropriate scale of $\lambda$
depends on the token length of the generation unit and is therefore chosen
per task domain.

\subsection{Prefix Buffer: Progressive Coverage}
\label{sec:prefix-buffer}

\emph{Motivation.} Early stopping saves computation but forfeits coverage
of the later turns. To recover that coverage without giving back the
speedup, we introduce the prefix buffer (PB), which caches high-quality
trajectory prefixes from prior rollouts and reuses them as context for
subsequent rollouts, so that the student's generation---and the training
that follows---concentrates on the weakest remaining turns.

\emph{Mechanism.} The PB operates in four steps during each training
iteration:

\begin{enumerate}
  \item \textbf{Quality decision.} If $\bar{l}_k > \alpha$, turn $k$ is deemed
        a ``correct turn,'' and its prefix is admitted as a reuse candidate.
        $\alpha$ is the prefix-quality threshold.
  \item \textbf{Weakest-turn localization.} Among all continuous correct
        prefix turns $1,\ldots,K_{\max}$ (where $K_{\max}$ is the length of the
        longest continuous correct prefix), locate the turn with the lowest
        teacher endorsement:
        \begin{equation}
          k^* = \arg\min\nolimits_{1 \le k \le K_{\max}} \bar{l}_k,
          \label{eq:weakest-turn}
        \end{equation}
        i.e.\ the weakest correct turn in the trajectory---the part most in
        need of training.
  \item \textbf{Prefix reuse.} In the next rollout of the same task, use the
        trajectory prefix up to turn $k^*{-}1$ as context, resample from turn
        $k^*$ onward, and apply the early-stopping rule
        (Eq.~\ref{eq:stop-rule}) to the newly generated portion.
  \item \textbf{Progressive coverage.} As the student gradually masters the
        earlier turns, their $\bar{l}_k$ values rise and $k^*$ shifts rightward
        naturally, so the training window slides toward the later stages of the
        trajectory without any hand-designed curriculum schedule.
\end{enumerate}

\emph{The $\alpha$ control surface.} Lowering $\alpha$ relaxes the quality
gate until, in the limit, filtering is disabled and every prefix becomes
reusable; raising it admits only prefixes with sufficient teacher
endorsement, guarding against prefix pollution at the cost of fewer reuse
opportunities. We study this trade-off in the ablation of
§\ref{sec:abl-alpha}.

In single-turn tasks (\emph{e.g.,} math), a single reasoning error does not necessarily imply that the
prefix is unusable (subsequent reasoning may still correct itself and return a
correct trajectory), so in that setting we disable $\alpha$ quality
filtering; in the multi-teacher setting, $\alpha$ filtering
plays a critical safety role (§\ref{sec:abl-alpha},
Figure~\ref{fig:alpha-abl}).

\subsection{Synergy and Overall Algorithm}
\label{sec:synergy}

\begin{figure}[t]
  \centering
  \resizebox{\columnwidth}{!}{%
  \begin{tikzpicture}[
    node distance=0.5cm and 0.6cm,                
    box/.style={rectangle, draw, rounded corners=2pt,
                minimum width=1.5cm, minimum height=0.6cm, 
                align=center, font=\scriptsize},           
    decision/.style={diamond, draw, aspect=1.8,
                     inner sep=0.5pt, align=center, font=\scriptsize},
    arrow/.style={-{Stealth[length=3.5pt]}, thick},        
    dashbox/.style={rectangle, draw, dashed, rounded corners=4pt,
                    inner sep=4pt},                         
    label/.style={font=\scriptsize\bfseries},
    ]

    \node[box, fill=blue!8] (task) {Task $x \in \mathcal{D}$};
    \node[decision, right=0.5cm of task, fill=yellow!10] (hasbuf)
      {Prior\\record?};
    \node[box, right=0.7cm of hasbuf, fill=green!8] (select)
      {Locate weakest\\$k^*{=}\arg\min \bar{l}_k$};
    \node[box, right=0.5cm of select, fill=green!8] (reuse)
      {Reuse prefix\\$y_{<k^*}$};

    \node[box, below=1.2cm of hasbuf, fill=orange!10] (rollout)
      {Student rollout\\(autoregressive)};
    \node[box, right=0.7cm of rollout, fill=orange!10] (teacher)
      {Teacher eval\\compute $\bar{l}_k$};
    \node[decision, right=0.5cm of teacher, fill=red!8] (ood)
      {$\sum \bar{l}_k$\\$< \lambda$?};

    \node[box, below=1.2cm of rollout, fill=purple!8] (train)
      {OPD loss $+$\\update $\theta$};
    \node[box, right=0.7cm of train, fill=blue!8] (record)
      {Record $\bar{l}_k$\\$\to \mathcal{H}(x)$};

    \draw[arrow] (task) -- (hasbuf);
    \draw[arrow] (hasbuf) -- node[above, font=\tiny] {yes} (select);
    \draw[arrow] (select) -- (reuse);
    \draw[arrow] (hasbuf) -- node[left, font=\tiny] {no} (rollout);
    \draw[arrow] (reuse.south) -- ++(0,-0.35) -| (rollout.north east);
    \draw[arrow] (rollout) -- (teacher);
    \draw[arrow] (teacher) -- (ood);

    \draw[arrow] (ood.south) -- ++(0,-0.6) -|
      node[above, near start, font=\tiny] {yes: stop} (train.north);

    \draw[arrow] (ood.north) -- ++(0,0.6) -|
      node[near start, above, font=\tiny] {no: continue} (rollout.north);

    \draw[arrow] (train) -- (record);

    \draw[arrow] (record.south) -- ++(0,-0.6) -|
      node[below, near start, font=\tiny] {next task} (task.south)
      -- (task.south);

    \begin{scope}[on background layer]
      \node[dashbox, fill=green!3, fit=(select)(reuse),
            label={[label, anchor=south]above:{Prefix buffer}}] {};
      \node[dashbox, fill=red!3, fit=(rollout)(teacher)(ood),
            label={[label, anchor=south]above:{Early stopping}}] {};
    \end{scope}
  \end{tikzpicture}%
  }
  \Description{Flowchart of the joint training framework: for each task, a
  prefix record is looked up, the weakest correct turn is located and its
  prefix reused; the student rolls out while the teacher evaluates each turn,
  and generation stops early once the cumulative teacher log-probability
  drops below a threshold; the OPD loss is then applied and the per-turn
  scores recorded for the next round.}
  \caption{Joint training framework of early stopping and prefix buffer.}
  \label{fig:framework}
\end{figure}
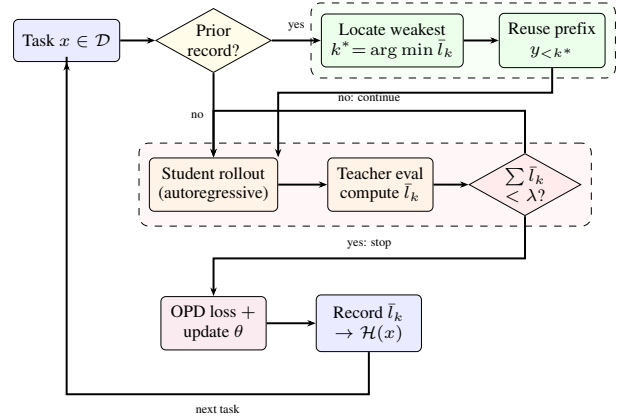

We refer to the combination of the two components as STRIDE: early stopping
answers ``when to stop,'' cutting the trajectory once it no longer provides
a reliable learning signal, and the prefix buffer answers ``where to
restart,'' refocusing training on the weakest remaining turns. The complete
training procedure is summarized in Algorithm~\ref{alg:opd-accel} and
illustrated in Figure~\ref{fig:framework}.

\begin{algorithm}[t]
\caption{STRIDE: accelerated OPD with adaptive early stopping and prefix
         buffer.}
\label{alg:opd-accel}
\SetKwInOut{Input}{Input}
\Input{Dataset $\mathcal{D}$, student $\pi_\theta$, teacher $\pi_T$,
       thresholds $\lambda$, $\alpha$}
Initialize per-task trajectory record $\mathcal{H} \leftarrow \emptyset$\;
\For{each training iteration}{
  \For{each task $x \in \mathcal{D}$}{
    \If{$\mathcal{H}(x)$ contains the prior rollout's trajectory record}{
      Take its per-turn teacher log-probs $\{\bar{l}_1,\ldots,\bar{l}_T\}$\;
      $K_{\max} \leftarrow \max\{K \mid \forall k \le K,\; \bar{l}_k > \alpha\}$\;
      $k^* \leftarrow \arg\min_{1 \le k \le K_{\max}} \bar{l}_k$\;
      Use the prefix up to turn $k^*{-}1$ as the context start,
      $K_0 \leftarrow k^*{-}1$\;
    }
    \Else{$K_0 \leftarrow 0$\;}
    \For{turn $k = K_0{+}1, K_0{+}2, \ldots$}{
      Student generates turn $k$; teacher computes $\bar{l}_k$\;
      \If{$\sum_{j=K_0{+}1}^{k} \bar{l}_j < \lambda$}{
        Early stop; \textbf{break}\;
      }
    }
    Record this trajectory's $\{\bar{l}_k\}$ to $\mathcal{H}(x)$\;
    Compute the OPD loss on the retained tokens (including the turn that
    triggered stopping) and update $\theta$\;
  }
}
\end{algorithm}

\section{Experiments}
\label{sec:experiments}

Our experiments are designed to answer three progressively broader questions:
(1) Is the proposed method effective in the core single-teacher agentic OPD
setting? (2) Can it generalize to cross-domain multi-teacher OPD? (3) Does its
effectiveness rely on the multi-turn structure of agentic tasks, or can it be
extrapolated to single-turn reasoning tasks?

\subsection{Experimental Setup}
\label{sec:setup}

\textbf{Models and infrastructure.} All experiments are implemented on the
veRL framework~\cite{sheng2024hybridflow}, with rollout handled by vLLM. The
three experiment configurations are:

\begin{itemize}
  \item \textbf{Single-teacher agentic} (§\ref{sec:single-teacher}): Student
        Qwen3-4B, teacher Qwen3-30B-A3B-Thinking-2507. The teacher doubles
        as the user simulator and the distillation-signal provider;
        multi-turn interaction follows the $\tau^2$-bench
        agentic specification~\cite{barres2025tau2bench}.
  \item \textbf{Multi-teacher agentic} (§\ref{sec:multi-teacher}): Student
        Qwen3-4B, with an additional Qwen3-8B scaling study, distilled jointly
        from a retail-domain and a telecom-domain expert teacher; an independent
        user simulator drives the dialogue but provides no distillation
        supervision, and each trajectory is scored by the expert teacher of its
        own domain.
  \item \textbf{Math reasoning} (§\ref{sec:math}): Student Qwen3-4B-Base,
        SFT-initialized on OpenThought3-8B reasoning traces; teacher Qwen3-8B.
        Single-turn chain-of-thought generation on dapo-math-17k.
\end{itemize}

  \textbf{Tasks and metrics.} Training tasks are synthesized on the
  $\tau^2$-bench environments following the data-generation pipeline of
  \citet{gao2026eigendata}. The single-teacher experiment is evaluated on
the retail domain with mean@16 and pass@16, the multi-teacher experiment on
retail and telecom with mean@4, and the math experiment on AIME 2024/2025
with mean@8; both agentic evaluations use the corrected $\tau^3$ release of
the task sets~\cite{sierra2026tau3fixes}. Further data and
evaluation details are deferred to Appendix~\ref{app:setup}.

\textbf{Compared methods.} \emph{Full-trajectory OPD (baseline):} full
rollout, all student-generated tokens participate in the loss, early stopping
and prefix buffer disabled. \emph{Fast~OPD}~\cite{zhang2026fastopd}: the
representative fixed-budget truncation method, which shortens the rollout to
a fixed budget. In the multi-turn agentic
experiments we evaluate its natural multi-turn extension with a fixed budget
of 5 interaction turns; in the math experiment we evaluate the original
token-budget variant (2048/4096 tokens). \emph{TCOD-F2B}~\cite{wang2026tcod}: temporal-curriculum
OPD, which controls the exposed trajectory depth by a manually designed
schedule and expands it over training (F2B variant). \emph{turnOPD}~\cite{zhou2026turnopd}:
turn-aware OPD for long-horizon agent training.
\emph{STRIDE w/o PB:} adaptive early stopping only ($\lambda$ active, prefix
buffer disabled). \emph{STRIDE:} the full combination of adaptive early
stopping and the prefix buffer, reusing a prefix whenever one is available.

\textbf{Threshold configuration.} The $\lambda$ values examined in each
setting are reported in the corresponding results tables; the prefix-quality
threshold $\alpha$ is fixed per setting (Appendix~\ref{app:setup}).

\subsection{Main Results: Single-Teacher Agentic OPD}
\label{sec:single-teacher}

\begin{table*}[t]
\centering
\caption{Single-teacher agentic OPD results ($\tau^2$-bench retail).
         Bold marks the best result in each column.}
\label{tab:single}
\small
\setlength{\tabcolsep}{10pt}
\begin{tabular}{@{}lcccc@{}}
\toprule
\textbf{Method} & \textbf{s/step} & \textbf{Speedup} & \textbf{mean@16} & \textbf{pass@16} \\
\midrule
\multicolumn{5}{l}{\emph{Reference}} \\
\textit{Teacher (30B)} & --- & --- & 0.461 & 0.847 \\
\midrule
\multicolumn{5}{l}{\emph{Baselines}} \\
Full OPD & 273.8 & 1.00$\times$ & 0.477 & 0.850 \\
Fast OPD~\cite{zhang2026fastopd} & 101.4 & 2.70$\times$ & 0.434 & 0.800 \\
TCOD-F2B~\cite{wang2026tcod} & 135.9 & 2.01$\times$ & 0.437 & 0.822 \\
turnOPD~\cite{zhou2026turnopd} & 167.4 & 1.64$\times$ & 0.434 & 0.801 \\
\midrule
\multicolumn{5}{l}{\emph{STRIDE w/o prefix buffer (ours)}} \\
$\lambda{=}{-}1$       & 61.8  & 4.43$\times$ & 0.368 & 0.732 \\
$\lambda{=}{-}1.5$     & 80.1  & 3.42$\times$ & 0.440 & 0.816 \\
$\lambda{=}{-}2$       & 95.7  & 2.86$\times$ & 0.451 & 0.823 \\
\midrule
\multicolumn{5}{l}{\emph{STRIDE w/ prefix buffer (ours)}} \\
$\lambda{=}{-}1$  & 73.3  & 3.73$\times$ & 0.475 & \textbf{0.859} \\
$\lambda{=}{-}1.5$ & 91.4  & 3.00$\times$ & 0.463 & 0.847 \\
\textbf{$\lambda{=}{-}2$} & 117.2 & 2.34$\times$ & \textbf{0.483} & 0.837 \\
\bottomrule
\end{tabular}
\end{table*}

\textbf{STRIDE matches or surpasses both the baseline and the teacher.}
With the prefix buffer, STRIDE stays on par with full-trajectory OPD at
every stopping threshold (Table~\ref{tab:single}; training curves in
Appendix~\ref{app:single-perf}): the best configuration attains the highest
mean@16 overall, exceeding both the full-OPD baseline and the 30B teacher,
and even the most aggressive threshold essentially matches the baseline's
mean@16 while achieving the best pass@16 of all methods. Notably, the student surpasses
its own teacher, so the acceleration does not come at the expense of this
capability.

\textbf{Comparison with fixed-budget and curriculum baselines.} Fast~OPD,
despite a $2.70\times$ speedup, falls below the baseline on both metrics: a
fixed turn budget cannot
distinguish high-quality trajectories from OOD ones---the former are cut off
prematurely while the latter are retained beyond the effective signal. The
curriculum-based TCOD-F2B and the turn-aware turnOPD close part of the gap
but remain clearly below the full-OPD baseline, at even lower speedups.
STRIDE dominates all three
on accuracy and speed simultaneously---even its most aggressive
configuration is both more accurate and faster than any of them. This
confirms that effective acceleration hinges on online OOD
detection rather than on a fixed budget or a hand-designed schedule.

\begin{figure}[t]
  \centering
  \includegraphics[width=0.85\columnwidth]{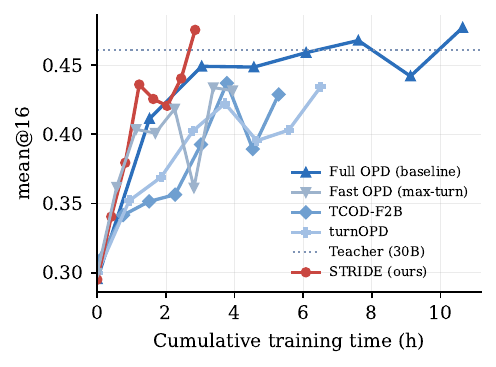}
  \Description{Line plot of mean at 16 accuracy against cumulative training
  hours for full OPD, STRIDE, and two curriculum baselines; STRIDE reaches
  baseline-level quality in about a quarter of the baseline's time.}
  \caption{Wall-clock efficiency: mean@16 vs.\ cumulative training hours.}
  \label{fig:single-wall}
\end{figure}

\textbf{Wall-clock efficiency.} Figure~\ref{fig:single-wall} shows mean@16
against cumulative training time. STRIDE with the prefix buffer, at its most
aggressive threshold, matches the baseline's final quality with about a
quarter of its time budget ($2.85$\,h vs.\ $10.65$\,h for the same 140
steps), while TCOD-F2B and turnOPD require considerably more time without
reaching baseline quality at any point; STRIDE thus dominates them on the
wall-clock axis as well.

\subsection{Main Results: Multi-Teacher Agentic OPD}
\label{sec:multi-teacher}

We extend the evaluation to cross-domain multi-teacher OPD on retail and
telecom (setup in §\ref{sec:setup}). This setting tests whether the
proposed acceleration remains effective under a more complex
student--teacher architecture (multiple domain-specialized teachers,
cross-domain joint training), and whether the prefix-quality threshold
$\alpha$ plays a measurable role when domain difficulty differs.

\begin{table*}[t]
\centering
\caption{Multi-teacher agentic OPD results ($\tau^2$-bench retail $+$ telecom,
         mean@4). Bold marks the best non-reference result in
         each domain column. Italic rows are references.}
\label{tab:multi}
\small
\setlength{\tabcolsep}{10pt}
\begin{tabular}{@{}lcccc@{}}
\toprule
\textbf{Method} & \textbf{s/step} & \textbf{Speedup} & \textbf{Retail} & \textbf{Telecom} \\
\midrule
\multicolumn{5}{l}{\emph{Reference: expert teachers}} \\
\textit{Retail} & --- & --- & 0.649 & 0.384 \\
\textit{Telecom} & --- & --- & 0.622 & 0.960 \\
\midrule
\multicolumn{5}{l}{\emph{Reference: single-domain full OPD}} \\
\textit{Retail} & 157.6 & --- & 0.586 & --- \\
\textit{Telecom} & 423.1 & --- & --- & 0.816 \\
\midrule
\multicolumn{5}{l}{\emph{Baselines}} \\
Full OPD & 221.5 & 1.00$\times$ & 0.590 & \textbf{0.853} \\
Fast OPD~\cite{zhang2026fastopd} & 47.8 & 4.63$\times$ & 0.548 & 0.783 \\
TCOD-F2B~\cite{wang2026tcod} & 71.3 & 3.11$\times$ & 0.561 & 0.752 \\
turnOPD~\cite{zhou2026turnopd} & 120.2 & 1.84$\times$ & 0.568 & 0.754 \\
\midrule
\multicolumn{5}{l}{\emph{STRIDE w/o prefix buffer (ours)}} \\
$\lambda{=}{-}1.5$   & 39.3 & 5.64$\times$ & 0.553 & 0.680 \\
$\lambda{=}{-}2$     & 43.8 & 5.06$\times$ & 0.533 & 0.739 \\
\midrule
\multicolumn{5}{l}{\emph{STRIDE w/ prefix buffer (ours)}} \\
$\lambda{=}{-}1.5$   & 37.3 & 5.94$\times$ & 0.550 & 0.818 \\
\textbf{$\lambda{=}{-}2$} & 49.1 & 4.51$\times$ & \textbf{0.599} & 0.807 \\
$\lambda{=}{-}3$     & 56.0 & 3.96$\times$ & 0.546 & 0.849 \\
\bottomrule
\end{tabular}
\end{table*}

The italic rows of Table~\ref{tab:multi} are calibration references excluded
from the speedup ranking: expert-teacher scores set domain-specific ceilings,
while single-domain full-OPD runs isolate the effect of cross-domain joint
training. We therefore compare each method only within its domain against the
corresponding teacher and baseline.

\begin{figure}[t]
  \centering
  \includegraphics[width=\columnwidth]{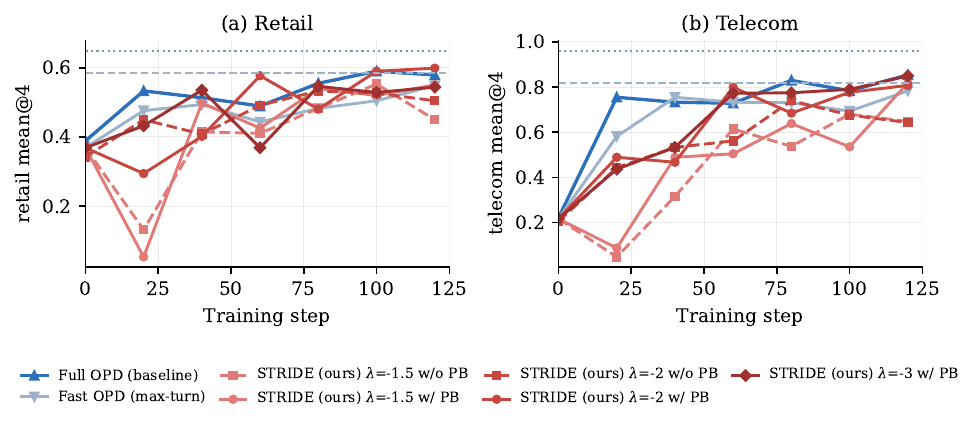}
  \Description{Two side-by-side line plots of retail and telecom mean at 4
  accuracy against training step for full OPD, fast OPD, and four STRIDE
  variants; prefix-buffer variants outperform their counterparts without the
  buffer on both domains.}
  \caption{Multi-teacher agentic OPD: retail (\textbf{a}) and telecom
           (\textbf{b}) mean@4 vs.\ training step. Solid lines use the
           prefix buffer ($\alpha=-0.8$); dashed lines are the
           early-stopping-only variants. The horizontal lines mark the
           single-domain baselines (gray dashed) and the domain expert
           teachers (purple dotted) of the respective domain.}
  \label{fig:multi-perf}
\end{figure}

\textbf{Gain of the accelerated methods over the cross-domain joint baseline.}
On retail, the best PB-based configuration surpasses the cross-domain joint
baseline at a substantial speedup, while the other PB configurations and the
w/o-PB variants remain below the baseline; the fixed-budget Fast~OPD and the
curriculum-based TCOD-F2B and turnOPD likewise fall short of the baseline
on both domains. This indicates that removing the
tokens whose teacher signal has degraded in OOD regions does not weaken the
effective supervision provided by the expert teachers---rather, it lets that
signal act more fully.

\textbf{Domain asymmetry.} On retail, the PB-based accelerated methods can
surpass the baseline; on the harder telecom domain, even the best
configuration still falls short of the baseline. A plausible contributor is
the difference in trajectory length: telecom tasks require substantially
more turns on average than retail ones---reflected in the larger test-time
turn budget of the multi-teacher setting (Appendix~\ref{app:setup})---while
the number of turns a single rollout can cover before the per-rollout
cumulative sum crosses $\lambda$ is roughly fixed, so each telecom rollout
covers a smaller fraction of the task and the training window advances
toward the later turns more slowly. Consistent with this, the most
conservative $\lambda=-3$ delivers the best accelerated telecom score
($0.849$), nearly matching the joint baseline ($0.853$), albeit at the cost
of retail accuracy ($0.546$).

\begin{table*}[t]
\centering
\caption{Multi-teacher agentic OPD results with a Qwen3-8B student
         ($\tau^2$-bench retail $+$ telecom, mean@4). Bold marks the best
         non-reference result in each domain column. Italic rows are
         references.}
\label{tab:multi-8b}
\small
\setlength{\tabcolsep}{10pt}
\begin{tabular}{@{}lcccc@{}}
\toprule
\textbf{Method} & \textbf{s/step} & \textbf{Speedup} & \textbf{Retail} & \textbf{Telecom} \\
\midrule
\multicolumn{5}{l}{\emph{Reference: expert teachers}} \\
\textit{Retail} & --- & --- & 0.649 & 0.384 \\
\textit{Telecom} & --- & --- & 0.622 & 0.960 \\
\midrule
\multicolumn{5}{l}{\emph{Baselines}} \\
Full OPD & 221.5 & 1.00$\times$ & 0.625 & 0.882 \\
Fast OPD~\cite{zhang2026fastopd} & 60.5 & 3.66$\times$ & 0.504 & 0.857 \\
TCOD-F2B~\cite{wang2026tcod} & 130.0 & 1.70$\times$ & 0.586 & 0.849 \\
turnOPD~\cite{zhou2026turnopd} & 113.4 & 1.95$\times$ & 0.548 & 0.671 \\
\midrule
\multicolumn{5}{l}{\emph{STRIDE w/o prefix buffer (ours)}} \\
$\lambda{=}{-}2$ & 56.1 & 3.95$\times$ & 0.610 & 0.807 \\
$\lambda{=}{-}3$ & 73.2 & 3.03$\times$ & 0.607 & 0.849 \\
\midrule
\multicolumn{5}{l}{\emph{STRIDE w/ prefix buffer (ours)}} \\
$\lambda{=}{-}2$ & 64.9 & 3.41$\times$ & 0.658 & 0.864 \\
\textbf{$\lambda{=}{-}3$} & 72.7 & 3.05$\times$ & \textbf{0.673} & \textbf{0.941} \\
\bottomrule
\end{tabular}
\end{table*}

\textbf{Scaling to a larger student.}
Table~\ref{tab:multi-8b} shows that STRIDE retains its qualitative advantage
with Qwen3-8B, achieving the strongest results on both domains among the
compared training methods while preserving a multi-fold speedup.

\subsection{Ablation Studies}
\label{sec:ablations}

We next isolate the contribution of each component of STRIDE: the prefix
buffer (§\ref{sec:abl-pb}), the stopping threshold $\lambda$
(§\ref{sec:abl-lambda}), and the prefix-quality threshold $\alpha$
(§\ref{sec:abl-alpha}).

\subsubsection{Effect of the Prefix Buffer}
\label{sec:abl-pb}

In both agentic settings the PB improves over the corresponding w/o-PB
variant at nearly every stopping threshold (Tables~\ref{tab:single}
and~\ref{tab:multi}), and the gain tends to grow with the aggressiveness of
stopping---consistent with the PB's role: the earlier a trajectory is
truncated, the larger the coverage gap the PB must bridge. The gain is also
consistently larger on the harder telecom domain than on the retail domain
of the same multi-teacher setting, indicating
that the harder the task, the more critical coverage compensation becomes.

\subsubsection{Effect of the Stopping Threshold $\lambda$}
\label{sec:abl-lambda}

The three $\lambda$ values with the PB define a clear speed--performance
tradeoff frontier (Table~\ref{tab:single}): a smaller $|\lambda|$ buys more
speedup at a negligible accuracy concession, while a larger $|\lambda|$
trades speedup for a robust accuracy lead. A user can thus select $\lambda$
directly by compute budget.

\begin{figure}[t]
  \centering
  \includegraphics[width=\columnwidth]{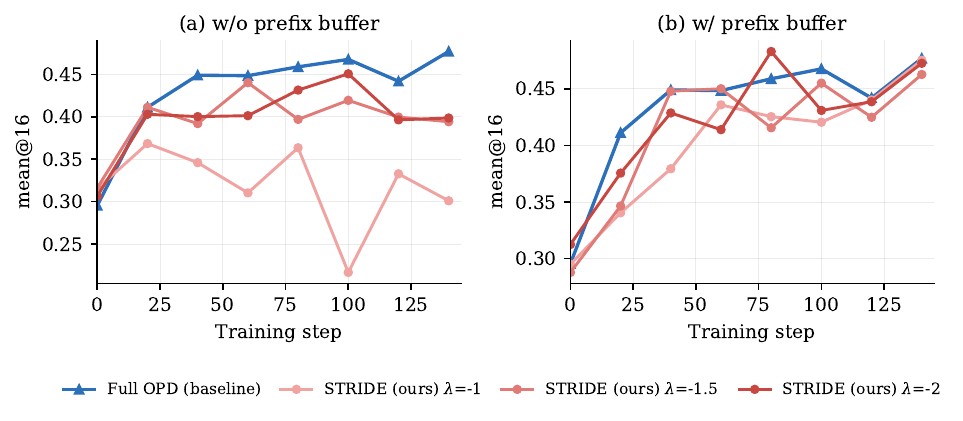}
  \Description{Two side-by-side line plots of mean at 16 accuracy against
  training step for three stopping thresholds, without and with the prefix
  buffer; without the buffer the truncated runs fall behind, with the buffer
  every threshold tracks the full-OPD baseline.}
  \caption{Single-teacher $\lambda$ ablation: mean@16 vs.\ training step
           without (\textbf{a}) and with (\textbf{b}) the prefix buffer.}
  \label{fig:single-lambda-abl}
\end{figure}

Figure~\ref{fig:single-lambda-abl} reports the full training curves behind
this frontier. Without the PB (panel a), all three thresholds fall behind
the baseline after about 60 steps---the coverage gap left by truncation
accumulates---and the most aggressive $\lambda=-1$ turns unstable; with the
PB (panel b), every threshold tracks the baseline throughout and the
differences between thresholds become small, indicating that the prefix
buffer makes the method robust to the choice of $\lambda$.

\subsubsection{Effect of the Prefix-Quality Threshold $\alpha$}
\label{sec:abl-alpha}

\begin{figure}[t]
  \centering
  \includegraphics[width=0.5\columnwidth]{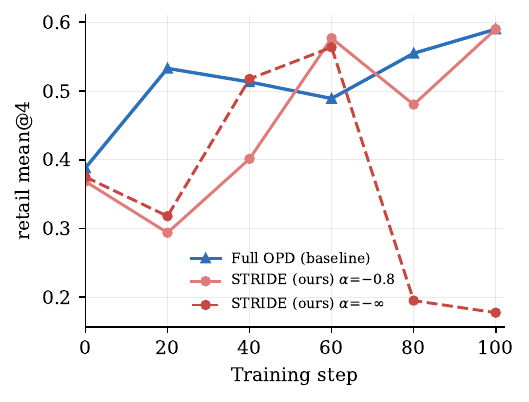}
  \Description{Line plot of retail mean at 4 accuracy over the first 100
  training steps comparing the prefix-quality filter settings; disabling the
  filter collapses training after briefly matching the filtered variant.}
  \caption{$\alpha$ ablation at $\lambda{=}{-}2$ on retail
           ($\tau^2$-bench), prefix reuse enabled, first 100 training steps.}
  \label{fig:alpha-abl}
\end{figure}

The prefix-quality threshold $\alpha$ controls whether low-endorsement
prefixes may be reused. Disabling this filter ($\alpha=-\infty$) admits
\emph{every} prefix for reuse, including those that embed the student's
earlier mistaken states: the next rollout then continues from a mistaken
state, early stopping triggers almost at the very start, and the effective
training length collapses. Holding $\lambda=-2$ fixed and contrasting the two
$\alpha$ choices on retail (Figure~\ref{fig:alpha-abl}) makes this failure
mode directly visible: the unfiltered variant matches the filtered one only
briefly before its curve collapses, while the filtered variant remains stable
throughout. We therefore default to $\alpha=-0.8$: $\alpha$ is not an
ornamental modulator but a key safety parameter that prevents prefix
pollution and safeguards reuse quality in difficult scenarios.

\subsection{Extension: Math Reasoning OPD}
\label{sec:math}

To test generality beyond multi-turn agents, we apply STRIDE to single-turn
math reasoning on AIME 2024/2025 at mean@8 (setup in §\ref{sec:setup}). Since
turn-level stopping is unavailable, we apply Eq.~\ref{eq:stop-rule}
incrementally and stop when the cumulative token-level teacher
log-probability falls below $\lambda$, enabling direct comparison with the
  2048/4096-token budgets of Fast~OPD~\cite{zhang2026fastopd}. Detailed results
  are reported in Appendix~\ref{app:math-results}. At a $5.10\times$ speedup,
  STRIDE improves AIME~2025 while trailing full OPD on AIME~2024; at a
  $3.08\times$ speedup, it improves AIME~2024 while remaining close to full OPD
  on AIME~2025. These results support teacher-signal OOD drift as a general
  mechanism beyond agentic tasks.

\section{Conclusion}
\label{sec:conclusion}

We presented STRIDE, an acceleration framework for on-policy distillation
that adapts the training budget to the online reliability of the teacher
signal. An analysis on $\tau^2$-bench shows that teacher supervision
concentrates in the within-turn prefix and that the cross-turn loss of
endorsement is locked to the student's first erroneous action; accordingly,
adaptive early stopping truncates rollouts at a cumulative
teacher-log-probability threshold, and the prefix buffer recycles verified
  prefixes from the weakest correct turn as a data-driven curriculum. Across
  single-teacher, cross-domain multi-teacher, and single-turn math-reasoning
  OPD, STRIDE provides competitive or better performance than full-trajectory
  OPD at multi-fold speedups; ablations identify the prefix buffer as the key
  coverage-compensation mechanism.

\noindent\textbf{Acknowledgments.} This work was supported by Ant Group
Research Intern Program.

\setlength{\bibsep}{0pt}
\bibliographystyle{assets/plainnat}
\bibliography{refs}

\beginappendix
\section{Experimental Setup Details}
\label{app:setup}

\paragraph{Hyperparameters.} The three experiments share the following
settings: AdamW optimizer, token-level OPD loss, maximum prompt length 24K (agentic) /
1K (math), and maximum response length 8K (agentic) / 30K (math).
Single-teacher: learning rate $10^{-5}$, batch size 24, student on 4\,GPUs
$+$ teacher on 4\,GPUs. Multi-teacher: learning rate $10^{-5}$, batch size 24,
student on 4\,GPUs $+$ the two domain-expert teachers and one user-simulator
instance together on 4\,GPUs. Math: learning rate $2\times10^{-6}$, batch size
48, student on 6\,GPUs $+$ teacher on 2\,GPUs. All experiments use FSDP with
parameter and optimizer-state offloading. The prefix-quality threshold is set
to $\alpha=-0.8$ for the agentic experiments; for the math task, $\alpha$
filtering is disabled ($\alpha=-\infty$).

\paragraph{Data and evaluation details.} For a fair comparison under a
common step budget, the single-teacher, multi-teacher, and math experiments
are truncated to the first 140, 120, and 160 training steps, respectively.
The Qwen3-8B multi-teacher scaling study otherwise follows the Qwen3-4B
configuration but uses the first 200 training steps because the larger student
converges more slowly and reaches a higher ceiling.
Each domain independently reports its best mean@4 within this budget;
speedups use unrounded mean training-step times.
The math student is SFT-initialized
on OpenThought3-8B reasoning traces for $500$ steps, and training uses at
most $2{,}880$ samples from dapo-math-17k. Evaluation runs every $20$
training steps, and all wall-clock measurements count only the per-step
training time, excluding evaluation. In the agentic Fast~OPD variant, the
5-turn budget applies only to training rollouts; the turn limit at test time
is $50$ in the single-teacher setting and $100$ in the multi-teacher
setting, the latter because telecom tasks require more turns.

\paragraph{Baseline configurations.} For turnOPD~\cite{zhou2026turnopd}, the
adaptive rollout budget is bounded by $H_{\min}{=}2$ and $H_{\max}{=}100$,
with EMA smoothing $\alpha_{\mathrm{ema}}{=}0.30$ on the control horizon, a
full-length probe every $8$ steps, $3$ warmup probe-only steps, coverage
quantile $p{=}0.80$, and a minimum of $8$ successful trajectories for
refreshing the coverage horizon; its turn-normalized loss weight is annealed
from $0$ (pure token-level) to $1$ (pure turn-normalized) over the first $10$
optimizer steps. For TCOD-F2B~\cite{wang2026tcod}, the number of supervised
turns grows as $k=\min(k_{\mathrm{start}}+\lfloor t/\eta\rfloor,\,
t_{\max})$ with $k_{\mathrm{start}}{=}1$, $\eta{=}10$, and $t_{\max}{=}50$.

\balance
\section{Math Reasoning Results}
\label{app:math-results}

\begin{table}[H]
\centering
\caption{Math OPD results (AIME 2024/2025, mean@8). Bold
         marks the best result in each column.}
\label{tab:math}
\scriptsize
\setlength{\tabcolsep}{2.5pt}
\begin{tabular}{@{}lcccc@{}}
\toprule
\textbf{Method} & \textbf{s/step} & \textbf{Speedup} & \textbf{AIME25} & \textbf{AIME24} \\
\midrule
\multicolumn{5}{l}{\emph{Baselines}} \\
Full OPD & 220.5 & 1.00$\times$ & 0.458 & 0.546 \\
Fast OPD 2048~\cite{zhang2026fastopd} & 22.4 & 9.83$\times$ & 0.408 & 0.554 \\
Fast OPD 4096~\cite{zhang2026fastopd} & 36.4 & 6.05$\times$ & 0.429 & 0.542 \\
\midrule
\multicolumn{5}{l}{\emph{STRIDE w/o prefix buffer (ours)}} \\
$\lambda{=}{-}1000$ & 41.6 & 5.30$\times$ & 0.429 & 0.542 \\
$\lambda{=}{-}2100$ & 63.7 & 3.46$\times$ & 0.429 & 0.533 \\
\midrule
\multicolumn{5}{l}{\emph{STRIDE w/ prefix buffer (ours)}} \\
\textbf{$\lambda{=}{-}1000$} & 43.2 & 5.10$\times$ & \textbf{0.467} & 0.517 \\
\textbf{$\lambda{=}{-}2100$} & 71.6 & 3.08$\times$ & 0.454 & \textbf{0.558} \\
\bottomrule
\end{tabular}
\end{table}

\section{Single-Teacher Training Curves}
\label{app:single-perf}

\begin{figure}[H]
  \centering
  \includegraphics[width=0.85\columnwidth]{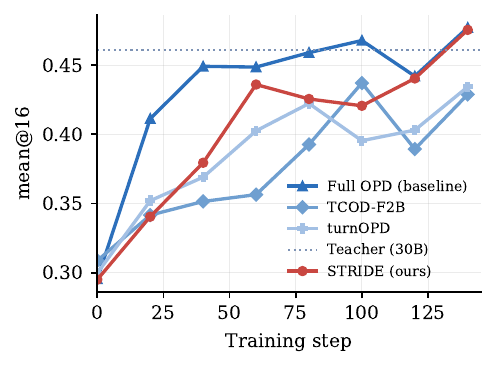}
  \Description{Line plot of mean at 16 accuracy against training step for
  full OPD, STRIDE, and two curriculum baselines; STRIDE tracks the baseline
  closely and ends above the 30B teacher.}
  \caption{Single-teacher agentic OPD: mean@16 vs.\ training step for
           full OPD, STRIDE ($\lambda=-1$ with the prefix buffer), and the
           curriculum baselines.}
  \label{fig:single-perf}
\end{figure}

Figure~\ref{fig:single-perf} reports the full mean@16 training curves of the
single-teacher agentic experiment (§\ref{sec:single-teacher}), complementing
the best-so-far numbers of Table~\ref{tab:single}.

\end{document}